\documentclass[10pt,twocolumn,letterpaper]{article}

\usepackage{cvpr}
\usepackage{times}
\usepackage{epsfig}
\usepackage{graphicx}
\usepackage{amsmath}
\usepackage{amssymb}
\usepackage{subfigure}
\usepackage{booktabs}
\usepackage{array}
\usepackage{lipsum}
\usepackage[abs]{overpic}
\usepackage{color}

\usepackage{float}
\usepackage{rotating}
\usepackage{makecell}

\usepackage[breaklinks=true,bookmarks=false]{hyperref}

\cvprfinalcopy 

\def\cvprPaperID{****} 

\begin{document}

\title{Real-Time Semantic Segmentation via Multiply Spatial Fusion Network}
\author{Haiyang Si$^{1, *}$ \hspace{22pt} Zhiqiang Zhang$^{2,}\thanks{The first two authors contribute equally to this work.}$ \hspace{22pt} Feifan Lv$^{1}$ \hspace{22pt} Gang Yu$^{2}$ \hspace{22pt} Feng Lu$^{1,3,4,}\thanks{Corresponding Author: Feng Lu (lufeng@buaa.edu.cn)}$\\
	\vspace{-10pt}
	\\{\normalsize$^{1}$State Key Laboratory of VR Technology and Systems, School of CSE, Beihang University, Beijing, China}
	\\{\normalsize$^{2}$Megvii Technology, Beijing, China \hspace{50pt} $^{3}$Peng Cheng Laboratory, Shenzhen, China}
	\\{\normalsize$^{4}$Beijing Advanced Innovation Center for Big Data-Based Precision Medicine, Beihang University, Beijing, China}
}

\maketitle

\begin{abstract}
	Real-time semantic segmentation plays a significant role in industry applications, such as autonomous driving, robotics and so on. It is a challenging task as both efficiency and performance need to be considered simultaneously.  To address such a complex task, this paper proposes an efficient CNN called Multiply Spatial Fusion Network (MSFNet) to achieve fast and accurate perception. The proposed MSFNet uses Class Boundary Supervision to process the relevant boundary information based on our proposed Multi-features Fusion Module which can obtain spatial information and enlarge receptive field. Therefore, the final upsampling of the feature maps of 1/8 original image size can achieve impressive results while maintaining a high speed. Experiments on Cityscapes and Camvid datasets show an obvious advantage of the proposed approach compared with the existing approaches. Specifically, it achieves 77.1\% Mean IOU on the Cityscapes test dataset with the speed of 41 FPS for a 1024$\times$2048 input, and 75.4\% Mean IOU with the speed of 91 FPS on the Camvid test dataset.
\end{abstract}

\section{Introduction}
Semantic segmentation aims to assign dense labels to each image pixel and is an essential task of computer vision. A variety of semantic segmentation techniques have been proposed to support different applications, such as autopilot, video surveillance and augmented reality. Existing methods mainly focus on improving the performance. However, achieving real-time performance with low latency is the most critical concerning for real applications. Therefore, how to maintain efficient inference speed and high accuracy becomes a challenging issue, especially for high-resolution images. 

\begin{figure}[t]
	\begin{center}
		\begin{overpic}[height=0.518\columnwidth]{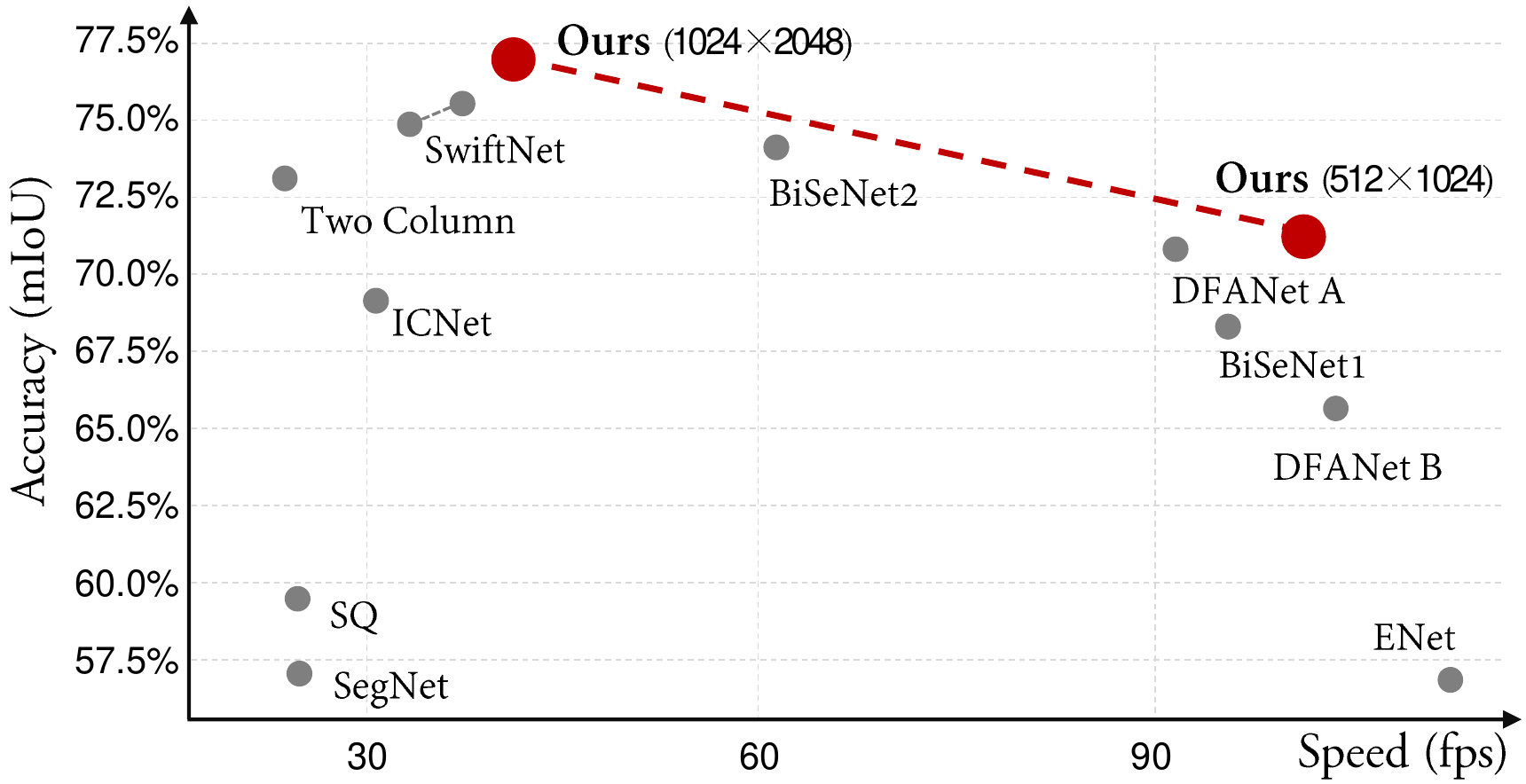}
		\end{overpic}
	\end{center}
	\vspace{-0.4cm}
	\caption{Inference speed and mIoU performance on the Cityscapes dataset. Our method significantly surpasses competitors in both accuracy and inference speed.}
	\label{fig:fig1}
	\vspace{-0.3cm}
\end{figure}

Recently, more and more researches~\cite{badrinarayanan2017segnet,zhao2018icnet,romera2017erfnet} focus on real-time semantic segmentation. Some methods~\cite{yu2018bisenet,li2019dfanet} accelerate the inference speed by reducing the input image's resolution, which will seriously lose spatial information, especially the edge-related one. Some other methods~\cite{badrinarayanan2017segnet,paszke2016enet} prune the number of feature channels to reduce computational cost.
However, these solutions will decrease the feature extraction capabilities of the network, which will result in a sharp drop in performance. 

\begin{figure*}
	\centering
	\includegraphics[height=4.8cm]{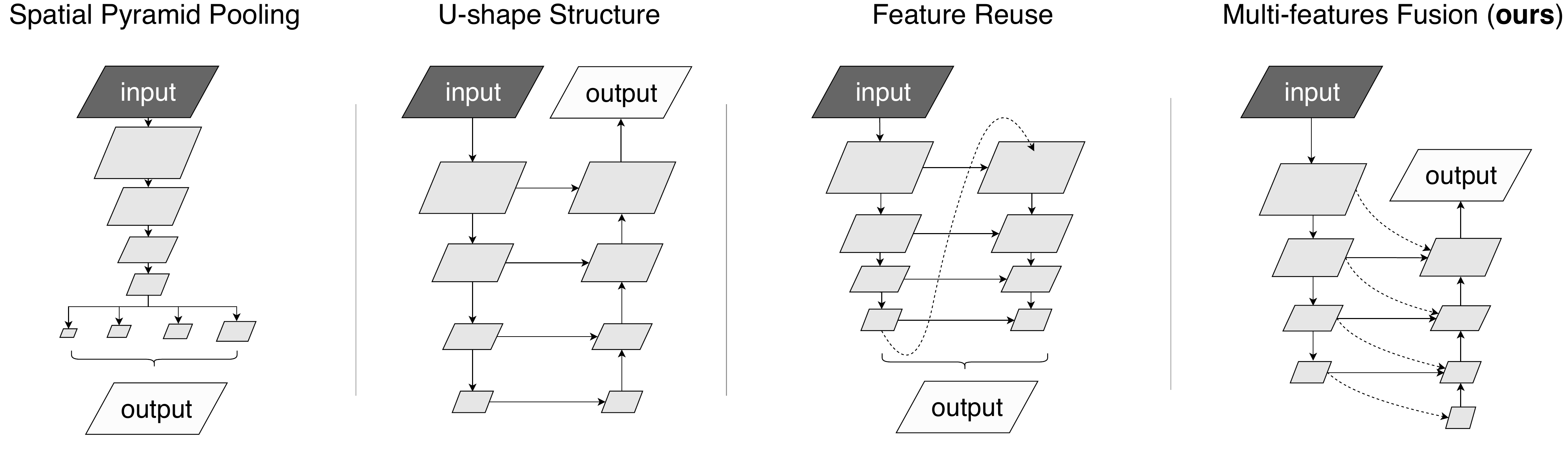}
	\caption{Network Architecture Comparison. As a comparison, the proposed architecture fuses multi-features to enlarge the receptive field and fully utilize spatial information. Besides, we directly upsample small feature maps (1/8  original image size) to speed up.}
	\label{fig:structure}
\end{figure*}

Another common solution for realizing real-time semantic segmentation is using shallow lightweight networks. However, these lightweight networks have obvious defects, as they are usually too shallow to achieve sufficient receptive field. These defects will make it difficult to preserve the spatial information of the object and result in performance degradation seriously. Moreover, the number of feature channels will be also reduced to improve the inference speed, which will limit the feature representation space. All these factors will limit the performance of existing networks for real-time semantic segmentation.

To address the dilemma of the real-time semantic segmentation, many network architectures have been proposed, as shown in Figure~\ref{fig:structure}. Spatial Pyramid Pooling (SPP)~\cite{chen2017rethinking,zhao2017pyramid} is a widely used structure to enlarge the receptive field. However, it will seriously increase computational cost and can't recover the loss of spatial information. U-shape structure~\cite{badrinarayanan2017segnet,ronneberger2015u} seems can alleviate above problems to some extent. However, a complete U-shape structure with large feature maps needs huge computational cost. 
Besides, it's difficult to achieve sufficient receptive field and perfectly recover the loss of spatial information only by merging feature maps. Another representative structure called Feature reuse~\cite{li2019dfanet} is beneficial to extract features and enlarge the receptive field.
It has the advantage of fewer parameters and faster inference speed. However, similar to SPP, it's incapable of recovering the loss of spatial information caused by downsampling.

By analyzing existing network architectures, we find that the key point is how to enlarge the receptive field and recover the loss of spatial information while maintaining a smaller computational cost. Based on this consideration, we propose an efficient lightweight network called Multiply Spatial Fusion Network (MSFNet), which can address above problems. The core component of MSFNet is the Multi-features Fusion Module(MFM), as shown in Figure~\ref{fig:structure}. It makes all different scale feature maps fuse with larger ones to enlarge the receptive field and recover more spatial information. Based on this special module, the final feature maps (1/8 original image size) will contain sufficient spatial information and significantly reduce the computational cost. Besides, we also propose Class Boundary Supervision (CBS) to avoid the loss of edge-related spatial information. 

Our proposed MSFNet achieves impressive results on Cityscapes~\cite{cordts2016cityscapes} and Camvid~\cite{brostow2008segmentation} benchmark datasets. More specifically, we obtain 77.1\% mIoU with 41 FPS and 71.3\% mIoU with 117 FPS on Cityscapes test set, and achieve 75.4\% mIoU with 91 FPS and 72.7\% mIoU with 160 FPS on Camvid test set, better than most of the state-of-the-art real-time segmentation methods. 

Our main contributions are summarized as follows:

\begin{itemize}
	\item We present a novel Multi-features Fusion Module (MFM) using the well-designed Spatial Aware Pooling (SAP) to enlarge the receptive field and recover the loss of spatial information while maintaining a small computational cost.
	\item We present a novel Class Boundary Supervision(CBS) to solve the loss of edge-related spatial information.
	\item Experiments on two benchmark datasets demonstrate that our method outperforms most of state-of-the-art methods in both accuracy and inference time.
\end{itemize}

\section{Related work}
{\bf Real-time Segmentation.} Many methods based on fully convolutional networks (FCNs)~\cite{chen2014semantic,long2015fully} have achieved high performances for semantic segmentation tasks. 
However, real-time semantic segmentation needs to consider both accuracy and inference speed. To reduce the computational cost, SegNet~\cite{badrinarayanan2017segnet} proposes a small network with a skip-connected method. ENet~\cite{paszke2016enet} presents a network with fewer downsamplings to pursue the ultimate rate.
ICNet~\cite{zhao2018icnet} uses multiple input sizes to capture objects of different sizes to improve the accuracy of real-time semantic segmentation. 
BiSeNet~\cite{yu2018bisenet} uses Spatial Path to recover spatial information and to implement real-time calculations. 
By redesigning the Xception network~\cite{chollet2017xception}, DFANet~\cite{li2019dfanet} uses Sub-network Aggregation and Sub-stage Aggregation to achieve extreme speed and maintain high accuracy.
Unlike existing network architectures, we carefully design a novel Multiply Spatial Fusion Network to enlarge the receptive field and recover the loss of spatial information while maintaining a small computational cost. Therefore, the proposed network is complementary to existing network architectures of real-time semantic segmentation.

\begin{figure*}
	\centering
	\includegraphics[height=8.5cm]{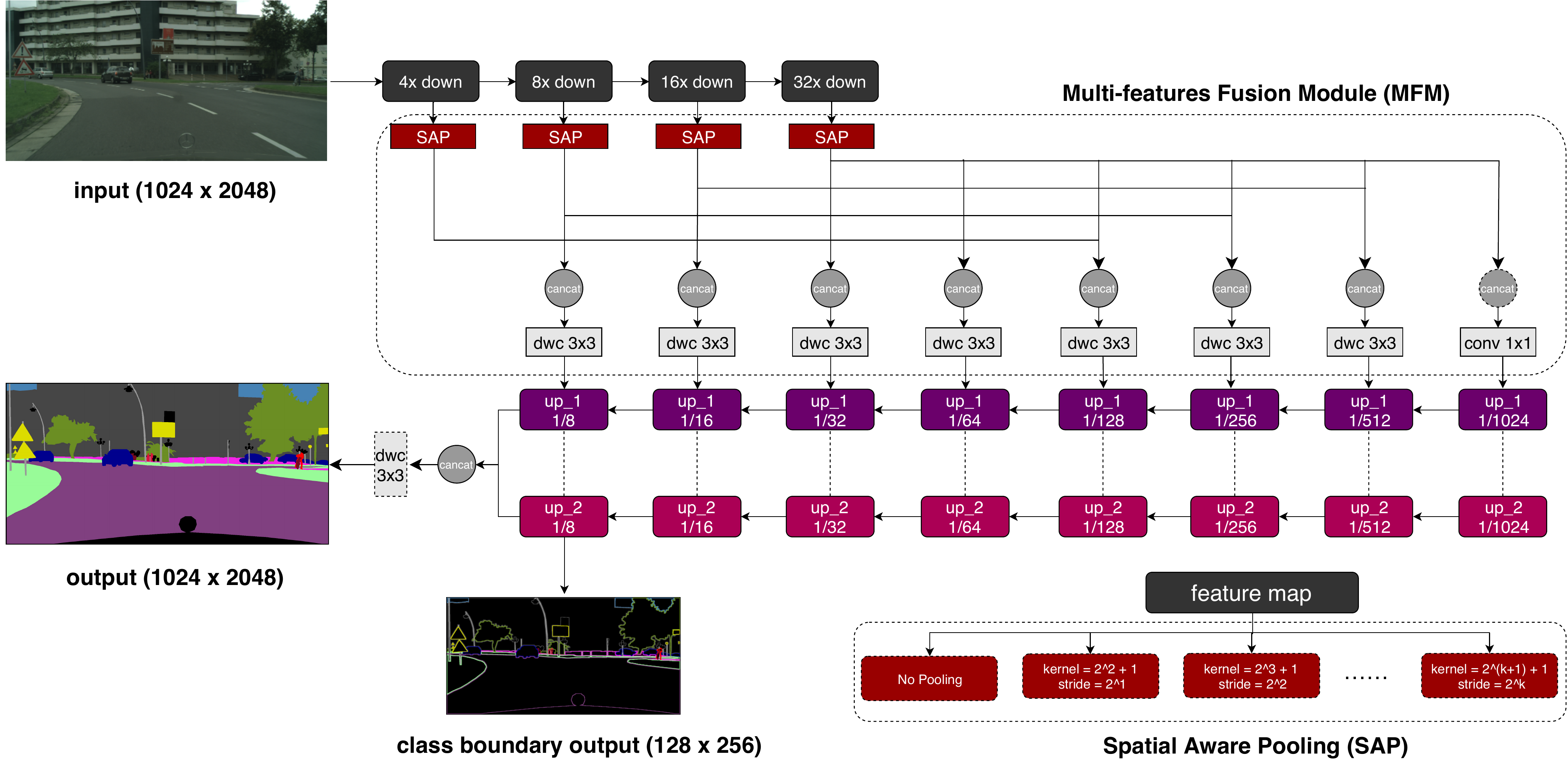}
	\vspace{0.08cm}
	\caption{Overview of our Multiply Spatial Fusion Network: multi-feature fusion module with spatial aware pooling and class boundary supervision. In this figure, ``dwc'' means depthwise separable convolution, ``N$\times$'' represents N times down-sampling operation, ``1/N'' means the feature size is 1/N of the input size.}
	\label{fig:pipeline}
\end{figure*}

{\bf Spatial Information.} Downsampling is a double-edged sword for semantic segmentation. On one hand, it enlarges the receptive field and enriches high-level features. On the other hand, it will result in a serious loss of spatial information. Paszke~\textit{et al.}~\cite{paszke2016enet} don't downsample to avoid losing spatial information, which will result in slow inference speed. Poude~\textit{et al.}~\cite{poudel2019fast} design a lightweight feature extraction network based on spatial information. Although its inference speed is very fast, its performance of extracting features is limited. Many networks use a U-shape structure to recover spatial information. U-Net~\cite{ronneberger2015u} uses the skip connection network to improve accuracy based on FCNs. However, the complete U-shape needs huge computational cost, especially for high resolutions images. Besides, it's difficult to achieve sufficient receptive field and perfectly recover the loss of spatial information only by merging feature maps. Considering these defects, we carefully design the network architecture to enlarge the receptive field and make the lightweight network more sensitive to spatial information.

{\bf Feature Fusion.} Feature fusion is widely used in semantic segmentation networks. As the increase of the network depth, the fusion and reuse of features show significant advantages. For instance, U-shape structure uses feature fusion to recover spatial information. RefineNet~\cite{lin2017refinenet} proposes a refine network module to finely fuse features.
Some other methods~\cite{yang2018denseaspp,yu2018deep,jegou2017one} use dense connections to improve the performance.
DFANet~\cite{li2019dfanet} proposes two feature fusion methods (Sub-network Aggregation and Sub-stage Aggregation) to enhance feature extraction capabilities. Our model uses feature fusion extensively to improve the interaction between different layers in terms of spatial information and semantic information, which improves the spatial sensitivity of the network obviously.

{\bf Boundary Supervision.} Many methods use boundary supervision to improve the accuracy of segmentation networks. Some methods~\cite{bertasius2016semantic,cheng2017fusionnet,lin2017refinenet} train a multi-task network at the same time, one of which is used for boundary detection. Most of these networks only classify the boundaries into one class and use boundary supervision at the loss functions, which means that they are just to use the boundaries for auxiliary supervision. Contrary to these methods, we use features extracted from the backbone to implement the boundary supervision with classes. Specially, the boundary feature maps are added to the network to supervise the spatial information of the boundary of objects.

\section{The Proposed Method}
In this section, we will elaborate on our proposed network. We first introduce our Multi-features Fusion Module with Spatial Aware Pooling in detail. Based on Multi-features Fusion Module, we then introduce the two branches proposed and highlight the Class Boundary Supervision. The whole architecture of Multiply Spatial Fusion Network (MSFNet) is shown in Figure~\ref{fig:pipeline}.

\subsection{Multi-features Fusion Module}
Existing real-time semantic segmentation networks typically use downsamplings to obtain high-level semantic information and to reduce computational cost. However, the spatial information in the high-level semantic layers will be seriously lost after multiple downsamplings. The objects' resolution in high-level feature maps is too small to accurately retain their shape. To address this problem, some methods utilize Spatial Pyramid Pooling (SPP) or Atrous Spatial Pyramid Pooling (ASPP) to capture sufficient receptive field. However, these special structures are usually used for enriching the high-level semantic information rather than the low-level spatial information.

{\bf Spatial Aware Pooling.} Based on the above analysis, we propose a novel structure called Spatial Aware Pooling (SAP), which following each residual block of the backbone. To extract rich features, we use some pooling with stride $s$ and kernel size $k=2s+1$. For high resolution (1024$\times$2048), we downsample the features of each residual block five times. More formally, the output of a residual block is denoted as $B_{i} \in R^{C\times\frac{H}{m_{i}}\times\frac{W}{m_{i}}}$, where height $H$ and width $W$ are the input size, $C$ is the number of channels and $m_{i}$ represents the stride. The pooling operation is defined as $\sigma ^{j}$ with stride $ s=2^{j}$ and kernel size $ k= 2^{j\times2+1}$, and we set $j \in \left[1,5\right]$. The outputs of the Spatial Aware Pooling can be defined as:
\begin{equation}
\centering
O_{i}^{j} = \left\{
\begin{array}{lcl}
B_{i}      &      & {j = 0}\\
\sigma^{j}(B_{i})     &      & {otherwise}\\

\end{array} \right.
\end{equation}
where $O_{i}^{j} \in R^{C\times\frac{H}{m_{i}\times2^{j}}\times\frac{W}{m_{i}\times2^{j}}}$ is the $j^{th}$ output of the Spatial Aware Pooling in the $i^{th}$ residual block.

In particular, $O_{1}^{1}$ is not used in upsampling because that the size of the feature maps of $O_{1}^{1}$ is 1/4 of the original image size, and our network structure is directly upsampled from the 1/8 of the original image size.

The sizable receptive field and accurate recovery of spatial information are both quite essential for semantic segmentation. In our model, the loss of spatial information can be recovered and the receptive field can be enlarged to a certain extent by our well-designed structure. Moreover, our proposed method has better recovery of spatial information at every receptive field level, and it greatly enhances the performance without increasing the cost of calculation.

{\bfseries Feature Fusion.} We aggregate the outputs with the same resolution of the Spatial Aware Pooling, and then fuse them by using a depthwise separable convolution layer with kernel $k=3$ that can reduce the cost of calculation because of the large number of channels after aggregation. So that not only can the features extracted by different layers in the backbone be merged to increase the mobility of information, but also the sensitivity of the semantic layers to spatial information can be enhanced.

\subsection{Class Boundary Supervision}
Many existing networks upsample to 1/4 of the original image size and then process bilinear interpolation to the original image size. We find that the 1/4 of the original image size is four times larger than the size of 1/8, which is 16 times larger than the size of 1/16. It means that in the case of the same number of channels, using 1/4 feature maps size requires 4 times cost of computation compared with using 1/8 feature maps size. Based on our proposed Multi-features Fusion Module, our upsampling branches can perform perfect segmentation results by upsampling from the feature maps of the 1/8 of the original image size.

We have noticed that the shallow layers in the encoder have rich spatial information. However, they can't fully recover the edge-related information because of its small final upsampling feature map size. In order to overcome the accuracy loss caused by the above problem, we propose an identical and independent multi-tasking upsampling decoder to achieve class boundary supervision.

Multi-features Fusion Module enriches high-quality features at every stage, which allows us to achieve satisfying segmentation results with a fast upsampling branch. In order to recover the edge spatial information and further improve the segmentation results, we propose two independent upsampling branches. During the upsampling process, the two upsampling branches do not transfer information to each other at all. As for boundary, we use the boundary of the ground-truth to supervise the segmentation task, which pays more attention to the edge contour.

Each upsampling stage has a different resolution. It has two inputs, one of which is the bilinear upsampling features of the previous stage, and the other is the output features of the Multi-features Fusion Module which has the same resolution as this stage. Finally, we fuse the two upsampling branches when the resolution is 1/8 of the original image size by using a depthwise separable convolution to get the final output which can slightly increase the speed.

\subsection{Network Architecture}
In conclusion, at first, Multi-features Fusion Module innovatively considers the improvement of both receptive field and spatial information simultaneously without adding extra computational cost due to its optimized network structure. It's a dense connection framework which is more efficient and is quite different from U-Shape. Secondly, we use an independent branch for edge-related information extraction, which can efficiently achieve class boundary supervision and correct the final semantic segmentation results.

In a word, our network structure is a typical encoder-decoder architecture. For real-time inference, we have to use a lightweight backbone as the encoder to extract features. And we use the Multi-features Fusion Module to support fast upsampling branches as the decoder to get better results. One thing to note, the class boundary supervision is an independent module, which can also be easily applied to other different network structures.

{\bfseries Backbone.}
In our network, the backbone is a lightweight Resnet-18 model, which is pre-trained on ImageNet. How to capture semantic context effectively is still a problem for semantic segmentation. Similarly, our Multi-features Fusion Module also needs rich context information. The Resnet-18 model has four different residual blocks, with each consisting of two 3$\times$3 convolutions and one skip connection. This kind of network design can better support the contextual requirements of our Multi-features Fusion Module. It can achieve real-time performance and guarantee high-quality feature extraction.

{\bfseries Loss Function.} 
In our network, we use an auxiliary loss function to supervise one of the upsampling branches to extract the edge-related spatial information. Besides, we use a principal loss function to supervise the output of the whole network. All the loss functions are standard cross-entropy (CE) loss, as shown in Equation 2. Moreover, we use a parameter $\lambda$ to balance the weight between the two loss function components, so that the network can better improve the performance of the segmentation, as Equation 3 presents. The parameter $\lambda$ is set to 1 in our network.
\begin{equation}
\centering
H_{y^{\prime}}(y) = \frac{1}{N}\sum_{i}L_{i} = -\frac{1}{N}\sum_{i}y_{i}^{\prime}log(y_{i})
\end{equation}
where $y$ denotes the prediction of the network and $y\prime$ denotes the ground-truth.
\begin{equation}
\centering
loss = H_{y_{s}^{\prime}}(y_{s}) + \lambda H_{y_{b}^{\prime}}(y_{b})
\end{equation}
where $y_{s}^{\prime}$ denotes ground-truth semantic labels and $y_{b}^{\prime}$ denotes ground-truth boundaries.

\section{Experiment}
As our model is designed to be highly efficient for high-resolution images, we evaluate performance on two challenging and representative datasets: Cityscapes and Camvid. We first introduce two datasets and the implementation details. Then, we analyze the effect of the proposed network and its components. Finally, we present the results of the accuracy and speed of the proposed network compared with other currently existing real-time semantic segmentation networks.

\subsection{Datasets}
{\bfseries Cityscapes.}
Cityscapes is a dataset which collects large urban street scenes from 50 different cities. It contains 5,000 finely annotated images and 19,998 coarsely annotated images with resolution up to 1024$\times$2048. According to the standard setting of Cityscapes, it divides the finely annotated images into 2975 images for training, 500 images for validation, and the remaining 1525 images for testing. In addition, it contains 30 classes, but only 19 of them are considered for training and evaluation. Our experiments use only finely annotated images.

{\bfseries Camvid.} Camvid is another well-known dataset for street scenes which is extracted from video sequences. It contains 701 annotated images, and according to the general method, 367 images are used for training, 101 for validation, and 233 for testing. This dataset contains 11 classes with resolutions up to 720$\times$960.

\subsection{Implementation Details}
The Adam optimizer~\cite{kingma2014adam} is adopted to train our model. Specifically, the batch size is set to 12 and the weight decay is set to $2.5\times 10^{-5}$. We use cosine attenuation~\cite{loshchilov2016sgdr} with initial learning rate to $10^{-4}$ and a minimum learning rate to $10^{-6}$. We train the model for 350 epochs on Cityscapes dataset, and Camvid is twice as many as Cityscapes. As for data augmentation, we use random horizontal flip and mean subtraction. We randomly use the parameters between $\left[0.5, 2\right]$ to transform the image to different scales, and then we randomly crop the resolution to 1024$\times$1024 on Cityscapes for training while the cropping resolution is 768$\times$1024 on Camvid.

\subsection{Network Structure Analysis}
In this section's experiments, we use the Cityscapes validation set for evaluation. For a fair comparison, we don't use any testing augmentation, such as multi-scale prediction or multi-model fusion. At the same time, in order to analyze the experiment more accurately, we use the mean of class-wise Intersection over Union (mIoU) as the evaluation standard.

{\bf Multi-features Fusion Module}
In this section, we explore the performance of the Multi-features Fusion Module in the proposed network. The Multi-model Fusion Module provides the necessary feature information for each stage in the upsampling branches.

As shown in the Table~\ref{tab:number of pooling}, as the number of pooling in each stage of the backbone increases from 0 to 2, the accuracy of the segmentation is greatly improved from 72.2\% to 75.3\%. Note that pooling times of 0 denotes an ordinary U-shape structure. When the number of pooling increases from 2 to 4, the accuracy of the segmentation decreases slightly. Finally, when the number of pooling increases from 4 to 5, the accuracy is greatly improved, and mIoU is as high as 77.2\%. We believe that the decline is within the normal fluctuation range. The reason for the performance improvement on the pooling times of 5 experiment is that the feature maps are quite small at this time, the model can fuse the global information, facilitate the propagation of gradient information and promote the model to better extract the information at different levels.

We also try to pool each stage in the backbone to the smallest feature map, but the result is worse, and the segmentation accuracy will drop from 77.2\% to 75.3\% compared to pooling times of 5 of each stage in the backbone. The reason for the decrease in the accuracy of the model is that the low-level layers in the backbone will bring some noise. Also, semantic information in the shallow feature maps is not rich, which may be detrimental to the extraction of high-level global semantic information.

\begin{table}[tbp]
	\centering
	\begin{tabular}{|c|c|}
		\hline
		Number of pooling & mIoU(\%) \\
		\hline
		Pooling $\times$0 & 72.2  \\
		Pooling $\times$1 & 72.6  \\
		Pooling $\times$2 & 75.3  \\
		Pooling $\times$3 & 75.1  \\
		Pooling $\times$4 & 74.9  \\
		Pooling $\times$5 & 77.2  \\
		Pooling to the end & 75.3  \\
		\hline
	\end{tabular}
	\vspace{0.3cm}
	\caption{Results on Cityscapes dataset with different numbers of pooling in each stage of the backbone, ``$\times$N'' means the number of pooling.}
	\label{tab:number of pooling}
\end{table}
\begin{table}[t]
	\centering
	\begin{tabular}{|c|c|}
		\hline
		Boundary mode & mIoU(\%) \\
		\hline
		0/1 boundary & 76.3  \\
		class boundary  & 77.2  \\
		\hline
	\end{tabular}
	\vspace{0.3cm}
	\caption{Results on Cityscapes dataset with 0/1 boundary and class boundary.}
	\label{tab:different boundary.}
\end{table}

\begin{table}[t]
	\centering
	\begin{tabular}{|c|c|c|}
		\hline
		Number of branches & Fusion methods & mIoU(\%) \\
		\hline
		1 & None & 75.3  \\
		1 & concat & 75.4  \\
		2 & None & 76.0  \\
		2 & concat & 77.2  \\
		\hline
	\end{tabular}
	\vspace{0.3cm}
	\caption{Results on Cityscapes dataset with with different number of branches and fusion methods.}
	\label{tab:different boundary.}
\end{table}

The pooling kernel size is $stride\times2+1$ which will increase the robustness of the model. Such a parameter setting manner enables each pixel in the feature maps to be captured by at least four windows, thereby enhancing the efficiency in feature fusion. Not surprisingly, the effect is worse when we replace the pooling with dilated convolutions. Although dilated convolutions can preserve spatial information while increasing the receptive field, our Spatial Aware Pooling does not need to maintain the original spatial resolution. As shown in Table~\ref{tab:pooling kernel size}, we can get 76.2\% mIoU result when the pooling kernel size is equal to stride but just 74.8\% when using 3$\times$3 dilated convolutions.

{\bfseries Class Boundary Supervision.}
In order to solve the loss of edge information, we propose Class Boundary Supervision(CBS) based on two completely independent upsampling branches, one of which is forcibly supervised to extract edge spatial information. In this section, we mainly explore the width of the boundaries of the ground-truth in the Class Boundary Supervision and the calculation of the boundary loss. Firstly, we define the boundary width parameter $\epsilon$. A pixel is a boundary pixel, if and only if there is at least one pixel within a distance of $\epsilon$ from it does not belong to the same class as the current pixel. Next, we need to define the ground-truth boundaries. A pixel's id is the original class if it is a boundary pixel, otherwise it is 0.

\begin{table}[t]
	\centering
	\begin{tabular}{|c|c|c|}
		\hline
		CBS size & width(px) & mIoU(\%) \\
		\hline
		1 & 1 & 75.5  \\
		1  & 3 & 75.9  \\
		1 & 5 & 75.8  \\
		1/8 & 1 & 77.2  \\
		1/8 & 3 & 76.1  \\
		1/8 & 5 & 75.4  \\
		\hline
	\end{tabular}
	\vspace{0.3cm}
	\caption{Results on Cityscapes dataset with different sizes of CBS branch output and different boundary width for boundary loss calculation. ``N'' of CBS size means N of the original image size.}
	\label{tab:boundary width}
\end{table}

\begin{table}[t]
	\centering
	\begin{tabular}{|c|c|}
		\hline
		Downsampling mode & mIoU(\%) \\
		\hline
		3$\times$3 dilated convolutions & 74.8  \\
		pooling($kernel = stride$)  & 76.1  \\
		pooling($kernel = stride\times2+1$)  & 77.2  \\
		\hline
	\end{tabular}
	\vspace{0.3cm}
	\caption{Results on Cityscapes dataset with different pooling kernel sizes and comparison with dilated convolutions.}
	\label{tab:pooling kernel size}
\end{table}

We use two methods to calculate the boundary loss. The first method will upsample the 1/8 feature maps to the original image size using bilinear upsampling, while the other method will calculate the loss directly in the 1/8 feature maps. As shown in Table~\ref{tab:boundary width}, bilinear upsampling to the original image size will result in poor performance. The reason is that bilinear upsampling will result in a continuous boundary, but the spatial boundary features at the 1/8 feature maps may be discontinuous, which will perturb the segmentation performance of the main branch, and thus cause drastic fluctuations. However, when the loss is calculated in the 1/8 feature map, large fluctuations caused by such discontinuous spatial features can be avoided. As can be seen from Table~\ref{tab:boundary width}, the highest segmentation accuracy can be obtained when the boundary width is 1. 

For comparison, we also perform different boundary-supervised experiments as shown in Table~\ref{tab:different boundary.}. We define the ground-truth with the boundary pixel id of 1 and another pixel id of 0 as 0/1 boundary. It can be seen that this strategy leads to a seriously reduction in the segmentation accuracy (from 77.2\% to 76.3\%). For the 0/1 boundary supervision method, it has a specific effect on the improvement of segmentation performance. However, as our boundary supervision is carried out in the final stage of the network, it cannot better identify classes.

\begin{table*}
	\centering
	\begin{tabular}{|c|c|c|c|c|}
		\hline
		Model & Input Size & FLOPS  & Frame(fps)  & mIoU(\%) \\
		\hline
		SegNet~\cite{badrinarayanan2017segnet} & 640$\times$360 & 286G & 16.7  & 57 \\
		ENet~\cite{paszke2016enet} & 640$\times$360 & 3.8G & 135.4  & 57 \\
		SQ~\cite{treml2016speeding} & 1024$\times$2048 & 270G & 16.7 &  59.8 \\
		ICNet~\cite{zhao2018icnet} & 1024$\times$2048 & 28.3G & 30.3  & 69.5 \\
		TwoColumn~\cite{wu2017real} & 512$\times$1024 & 57.2G & 14.7 & 72.9 \\
		BiSeNet1~\cite{yu2018bisenet} & 768$\times$1536 & 14.8G & 105.8  & 68.4 \\
		BiSeNet2~\cite{yu2018bisenet} & 768$\times$1536 & 55.3G & 65.5  & 74.7 \\
		DFANet A~\cite{li2019dfanet} & 1024$\times$1024 & 3.4G & 100  & 71.3 \\
		DFANet B~\cite{li2019dfanet} & 1024$\times$1024 & 2.1G & 120  & 67.1 \\
		SwiftNet pyr~\cite{orsic2019defense} & 1024$\times$2048 & 114.0G & 34.0  & 75.1 \\
		SwiftNet~\cite{orsic2019defense} & 1024$\times$2048 & 104.0G & 39.9  & 75.5 \\
		\hline
		Ours (without CBS) & 1024$\times$2048 & 85.2G & 47  & 75.4 \\
		Ours (with CBS) & 512$\times$1024 & 24.2G & 117 &  71.3 \\
		Ours (with CBS) & 1024$\times$2048 & 96.8G & 41 &  \textbf{77.1} \\
		\hline
	\end{tabular}
	\vspace{0.3cm}
	\caption{Accuracy and speed analysis on Cityscapes test dataset. "-" indicates that the corresponding result is not provided.}
	\label{tab:headings}
\end{table*}

Finally, We perform an ablation experiment on the number of upsampling branches and fusion methods for the network with CBS modules. It can be seen that the performance is poor regardless of fusion when an upsampling branch is used. This is because that the upsampling branch of our network is a simple decoder that can't learn multi-tasks at the same time, and the noise caused by multi-tasks can affect the semantic segmentation performance. When two branches are used to eliminate noise, the performance of semantic segmentation is greatly improved, and 76.0\% of mIoU can be obtained even without fusion.

\begin{table}
	\centering
	\begin{tabular}{|c|c|}
		\hline
		Model & mIoU(\%) \\
		\hline
		U-shape-8s & 71.1  \\
		U-shape-8s + MFM  & 76.0  \\
		U-shape-8s + MFM + CBS  & 77.2  \\
		\hline
	\end{tabular}
	\vspace{0.3cm}
	\caption{Results on Cityscapes dataset with different modules}
	\label{tab:different module}
\end{table}

{\bf The Whole Network.} Finally, we focus on the whole results of our proposed network. As shown in Table~\ref{tab:different module}, based on ``U-shape-8s" which the final upsampling is also performed from the feature maps of the 1/8 original size, our Multi-feature Fusion Module can achieve huge improvement and improve its accuracy from 71.1\% to 76.0\%. In order to solve the loss of the boundary information, our proposed Class Boundary Supervision based on two upsampling branches can improve the segmentation performance of the model again, and from 76.0\% to 77.2\%.

\subsection{Speed and Accuracy Comparison}
\begin{figure*}
	\centering
	\includegraphics[height=7.5cm]{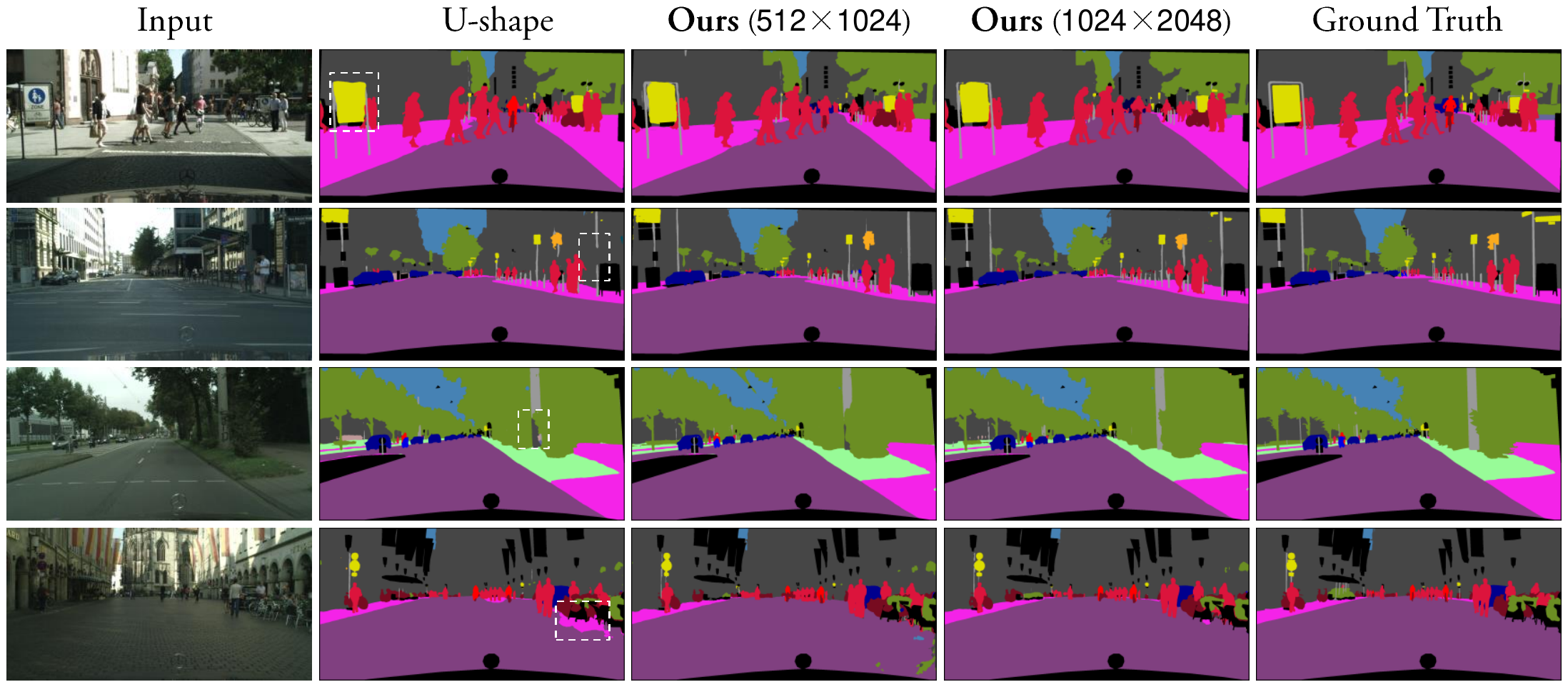}
	\vspace{0.3cm}
	\caption{Example results of the different methods on Cityscapes dataset. The first line is input images, and Lines 2-4 display the results of U-shape network, MSFNet with 512$\times$1024 input and MSFNet with 1024$\times$2048 input. The final line is the ground-truth.}
	\label{fig:vis}
\end{figure*}

The comparison of the accuracy and speed of our model is shown in Table~\ref{tab:headings}. We report the average time of 500 predictions using images of 1024$\times$2048 resolution. Our experimental environment is a single GTX 2080 Ti GPU on a virtual machine. Similar to Swiftnet~\cite{orsic2019defense}, we exclude batch normalization layers when inference, because they can be merged with the previous convolution in real-time semantic segmentation. Benefit from our model's bilinear upsampling of 1/8 original size, following BiSeNet~\cite{yu2018bisenet}, writing them out of the GPU memory and then upsampling could save a small amount of time. Some visual results of the proposed MSFNet are shown in Figure~\ref{fig:vis}. We can achieve high performance of semantic segmentation on Cityscapes using our proposed Multiply Spatial Fusion Network.

As depicted in Table~\ref{tab:headings} and Figure~\ref{fig:fig1}, our solution outperforms all existing real-time semantic segmentation methods, and the speed remains comparable. Without Multiple Upsampling Branches, our model is still able to achieve the state-of-the-art performance, specifically it can achieve 75.4\% mIoU with 47 FPS on Cityscapes test set. Our final model eliminates some effects of edge variation caused by downsampling. Finally it can achieve 77.1\% mIoU with 41 FPS for a 1024$\times$2048 input and 71.3\% mIoU with 117 FPS for a 512$\times$1024 input on Cityscapes test set. Note that for 512$\times$1024 inputs we only downsample 4 times.

Some networks use the backbone with a large number of depthwise separable convolution layers. Therefore, it naturally has quite small FLOPs. However, depthwise separable convolutions are not directly supported in GPU firmware (like the cuDNN library, etc.), although these networks have small FLOPs, our experiments show that MSFNet can maintain the same level of accuracy and speed.

\begin{table}[]
	\centering
	\begin{tabular}{|c|c|c|}
		\hline
		Model & Frame(fps) & mIoU(\%) \\
		\hline
		DPN~\cite{yu2018learning} & 1.2 & 60.1  \\
		DeepLab~\cite{chen2017deeplab} & 4.9 & 61.6  \\
		ENet~\cite{paszke2016enet} & - & 51.3  \\
		ICNet~\cite{zhao2018icnet} & 27.8 & 67.1  \\
		BiSeNet1~\cite{yu2018bisenet} & - & 65.6  \\
		BiSeNet2~\cite{yu2018bisenet} & - & 68.7  \\
		DFANet A~\cite{li2019dfanet}  & 120 & 64.7  \\
		DFANet B~\cite{li2019dfanet} & 160 & 59.3  \\
		SwiftNet pyr~\cite{orsic2019defense} & - & 72.85  \\
		SwiftNet~\cite{orsic2019defense} & - & 73.86  \\
		\hline
		Ours (without CBS) & 101 & \textbf{73.9}  \\
		Ours$^{*}$ (with CBS) & \textbf{160} & 72.7  \\
		Ours (with CBS) & 91 & \textbf{75.4}  \\
		\hline
	\end{tabular}
	\vspace{0.3cm}
	\caption{Accuracy and speed analysis on CamVid test dataset. "-" indicates that the corresponding result is not provided by the methods. Ours$^{*}$ is 512$\times$768 input and others are 768$\times$1024 input.}
	\label{tab:cavid}
\end{table}

Based on the proposed two novel structures, our network can achieve state-of-the-art results, which get the highest precision in the current real-time semantic segmentation field, while other non-real-time methods all consume more than 1 second even more.

\subsection{ Results on Other Dataset.}
To validate the generality of our method, we also experimented on the Camvid dataset which has a resolution of 720$\times$960. To better fit our model, we resize the images to 768$\times$1024 for training and testing. Compared to the previous 5 times of downsampling in Spatial Aware Pooling, we changed it to 3 to better adapt to this resolution of the images, which shows the stability of our model that can work at different resolutions. According to Table~\ref{tab:cavid}, our model is capable of achieving the mIoU of 75.4\% and the FPS of 91. This result is beyond the accuracy of all current existing real-time semantic segmentation networks, and the speed is also comparable.

To further verify the superiority of our proposed method, we also conducted multiple experiments on the Camvid dataset. By reducing the resolution of the input, it achieves 72.65\% mIoU and 160 FPS while the input resolution is 512$\times$768. This result is similar to the Cityscapes dataset, which shows that our network can achieve excellent performance at different resolutions. Finally, we conducted an experiment without CBS and get 73.90\% mIoU, which is still state-of-the-art. The large improvement by adding CBS shows the effectiveness of the CBS on different datasets.

\section{Conclusion}
In this paper, we propose a novel Multi-feature Fusion Module based on Spatial Aware Pooling, which greatly improves the performance of real-time semantic segmentation. Based on it, we propose Class Boundary Supervision in order to recover the loss of edge-related spatial information. Finally, we verify the effectiveness of our approach on Cityscapes and Camvid benchmark datasets. The result obviously shows that our method outperforms existing methods by a large merge in both speed and accuracy.

{\small
	\bibliographystyle{ieee_fullname}
	\bibliography{egbib}
}

\end{document}